\documentclass[10pt,times]{article}

\usepackage{graphicx}
\usepackage[utf8]{inputenc}
\usepackage{comment}
\usepackage{hyperref}

\newcommand{\bb}{\textbf{b}}
\newcommand{\bx}{\textbf{x}}
\newcommand{\bL}{\textbf{L}}
\newcommand{\bA}{\textbf{A}}
\newcommand{\bW}{\textbf{W}}
\newcommand{\bmu}{\mbox{\boldmath$\mu$}}
\newcommand{\bU}{\textbf{U}}
\newcommand{\bSigma}{\mbox{\boldmath$\Sigma$}}
\newcommand{\bLambda}{\mbox{\boldmath$\Lambda$}}

\newcommand{\aaron}[1]{{{\bf[AH: #1]}}}

\title{Preserving Color in Neural Artistic Style Transfer}
\author{Leon A. Gatys \\ U.~T\"{u}bingen \and Matthias Bethge \\ U.~T\"{u}bingen \and Aaron Hertzmann \\ Adobe Research \and Eli Shechtman \\ Adobe Research}
\date{June 2016}

\begin{document}

\maketitle

\begin{figure}[h]
\includegraphics{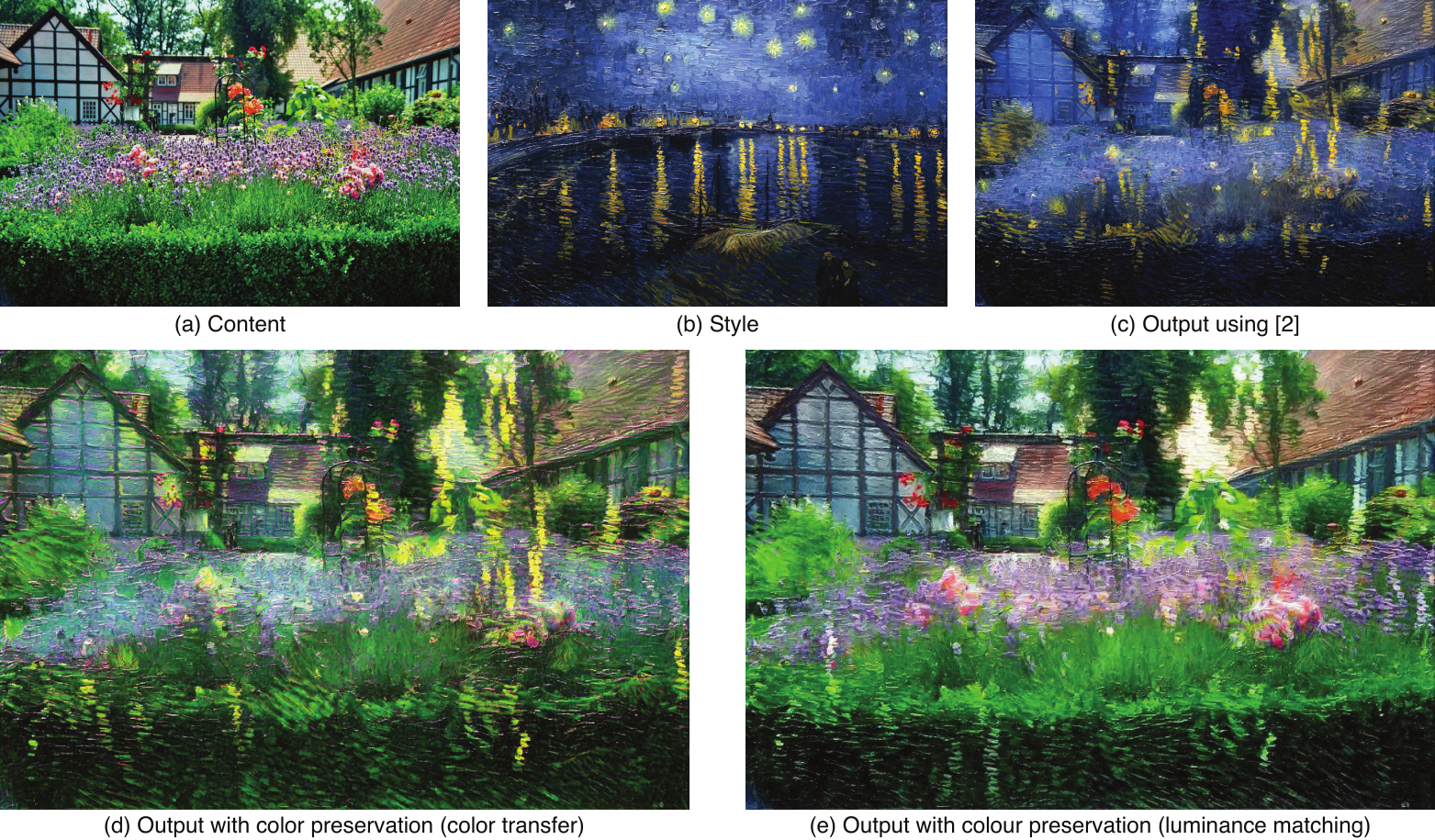}
\caption{\label{fig:teaser}
Example using a style dominated by brushstrokes.
\textbf{(a)} Input photograph.
\textbf{(b)} Painting \textit{Starry night over the Rhone} by Vincent van Gogh.
\textbf{(c)} Transformed content image, using original neural style transfer algorithm \cite{gatys}. The color scheme is copied from the painting.
\textbf{(d)} Transformed content image, using color transfer to preserve colors.
\textbf{(e)} Transformed content image, using style transfer in luminance domain to preserve colors.
}
\end{figure}

\begin{abstract}
    This note presents an extension to the neural artistic style transfer algorithm \cite{gatys}. The original algorithm transforms an image to have the style of another given image. For example, a photograph can be transformed to have the style of a famous painting. Here we address a potential shortcoming of the original method: the algorithm transfers the colors of the original painting, which can alter the appearance of the scene in undesirable ways. We describe simple linear methods for transferring style while preserving colors.
\end{abstract}

\section{Introduction}

\begin{figure}[h]
\includegraphics{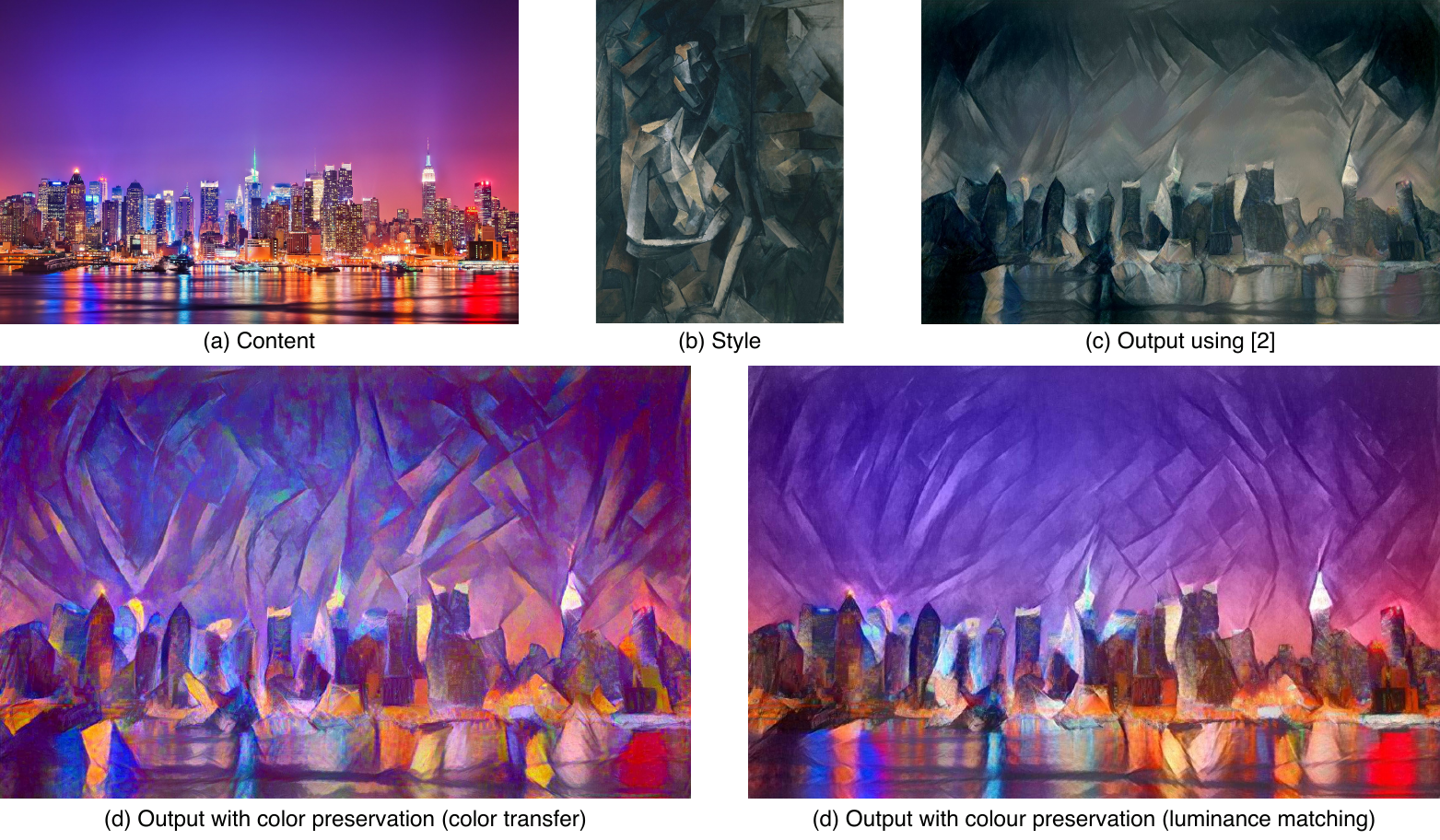}
\caption{\label{fig:teaser_NY}
Example using a style dominated by geometric shapes.
\textbf{(a)} Input photograph.
\textbf{(b)} Painting \textit{Femme nue asisse} by Pablo Picasso.
\textbf{(c)} Transformed content image, using original neural style transfer algorithm \cite{gatys}. The color scheme is copied from the painting.
\textbf{(d)} Transformed content image, using color transfer to preserve colors.
\textbf{(e)} Transformed content image, using style transfer in luminance domain to preserve colors.
}
\end{figure}

The recent neural artistic style algorithm \cite{gatys} takes as input two images --- a style image and a content image --- and outputs a new image depicting the objects of the content image in the style of the other image. For example, Figure \ref{fig:teaser} shows a photograph of a farmhouse garden as the content image, and a painting by Vincent van Gogh as the style image. Applying the algorithm produces a new painting of the farmhouse in the style of the van Gogh painting.
This method works by matching statistics on the feature responses in a  Convolutional Neural Network trained on object recognition; for a detailed description of the algorithm see \cite{gatys}.
As illustrated in Figures \ref{fig:teaser} and \ref{fig:teaser_NY},  the output reproduces the style of brushstrokes, geometric shapes, and painterly structures exhibited in the style image. However, it also copies the color distribution of the style image which might be undesirable in many cases (Fig.~\ref{fig:teaser}(c), \ref{fig:teaser_NY}(c)).
%A goal of this work is to produce an image similar to what the original artist might have painted, if confronted with this scene. 
For example, the painting of the farmhouse has the colors of the original van Gogh painting (Fig.~\ref{fig:teaser}(c)), whereas one might prefer the output painting to preserve the colors of the farmhouse photograph. In particular, one might also imagine that the artist would have used the colors of the scene if they were to paint the farmhouse.

This note presents two simple methods to preserve the colors of the source image during neural style transfer---in other words, to transfer the style without transferring the colors. 
 %Image Analogies \cite{image-analogies, hertzmann-phd}, a previous style-transfer algorithm, though we find that different variants work best for neural transfer. 
%\leon{which approach was used in Image Analogies? the color transfer or the luminance matching - I always though the luminance matching?}
We compare two different approaches to color preservation: color histogram matching and luminance-only transfer
(Fig.~\ref{fig:teaser}(d,e), \ref{fig:teaser_NY}(d,e)). We compare these methods and discuss their advantages and disadvantages.

\section{Approach \#1: Color histogram matching}

The first method we present works as follows.
%transfers the color distribution from the content image onto the style image before the style transfer.  The original neural style transfer algorithm is then applied as normally.
Given the style image ($S$), and the content image ($C$),
the style image's colors are tranformed to match the colors of the content image. 
This produces a new style image $S'$ that replaces $S$ as input to the neural style transfer algorithm. The algorithm is otherwise unchanged.

The one choice to be made is the color transfer procedure.
There are many color transformation algorithms to choose from; see \cite{CGF:CGF12671} for a survey. 
Here we use linear methods, which are simple and very effective for color style transfer. %, to transform the colors of the style image to match the second-order statistics of the content image's colors. 

In particular, let $\bx_i = (R,G,B)^T$ be a pixel of an image. 
Each pixel is transformed as:
\begin{equation}
\bx_{S'} \leftarrow \bA \bx_S + \bb
\end{equation}
where $\bA$ is a $3\times3$ matrix and $\bb$ is a 3-vector. 

We choose this transformation so that the mean and covariance of the RGB values in the new style image $S'$ match those of $C'$. 
In other words, let $\bmu_S$ and $\bmu_C$ be the mean colors of the style and content images, and $\bSigma_S$ and $\bSigma_C$ be the pixel covariances. 
The mean and covariance of the color pixels is given by $\bmu = \sum_i \bx_i / N$ and $\bSigma = \sum_i (\bx_i - \bmu) (\bx_i - \bmu)^T / N$.  We want to choose $\bA$ and $\bb$ to be satisfy $\bmu_{S'} = \bmu_C$ and $\bSigma_{S'} = \bSigma_C$. 
\begin{comment}
\begin{eqnarray}
\bmu &= &\frac{1}{N} \sum_i \bx_i / N \\
\bSigma& =& \frac{1}{N} \sum_i (\bx_i - \bmu) (\bx_i - \bmu)^T
\end{eqnarray}
We choose $\bA$ and $\bb$ to satisfy
\begin{eqnarray}
\bmu_{S'} &=& \bmu_{C} \\
\bSigma_{S'} &=& \bSigma_C
\end{comment}
These are satisfied by the constraints:
\begin{eqnarray}
\bb &=& \bmu_C - \bA \bmu_S \\
\bA \bSigma_S \bA^T &=& \bSigma_C
\end{eqnarray}
There remains a family of solutions for $\bA$ that satisfies these constraints. 

We consider two variants. The first variant uses the Cholesky decomposition: 
\begin{eqnarray}
\bA_{\mathrm{chol}} = \bL_C \bL_S^{-1}
\end{eqnarray}
where $\bSigma = \bL \bL^T$ is the Cholesky decomposition of $\bSigma$.

The second variant is a 3D color matching formulations explored in Image Analogies \cite{hertzmann-phd} (Appendix B).  First, let the eigenvalue decomposition of a covariance matrix be $\bSigma = \bU \bLambda \bU^T$. Then, we define a matrix square-root as: $\bSigma^{1/2} = \bU \bLambda^{1/2} \bU^T$. Then, the transformation is given by
\begin{eqnarray}
\bA_{IA} = \bSigma_C^{1/2} \bSigma_S^{-1/2}
\end{eqnarray}

\begin{figure}
\includegraphics{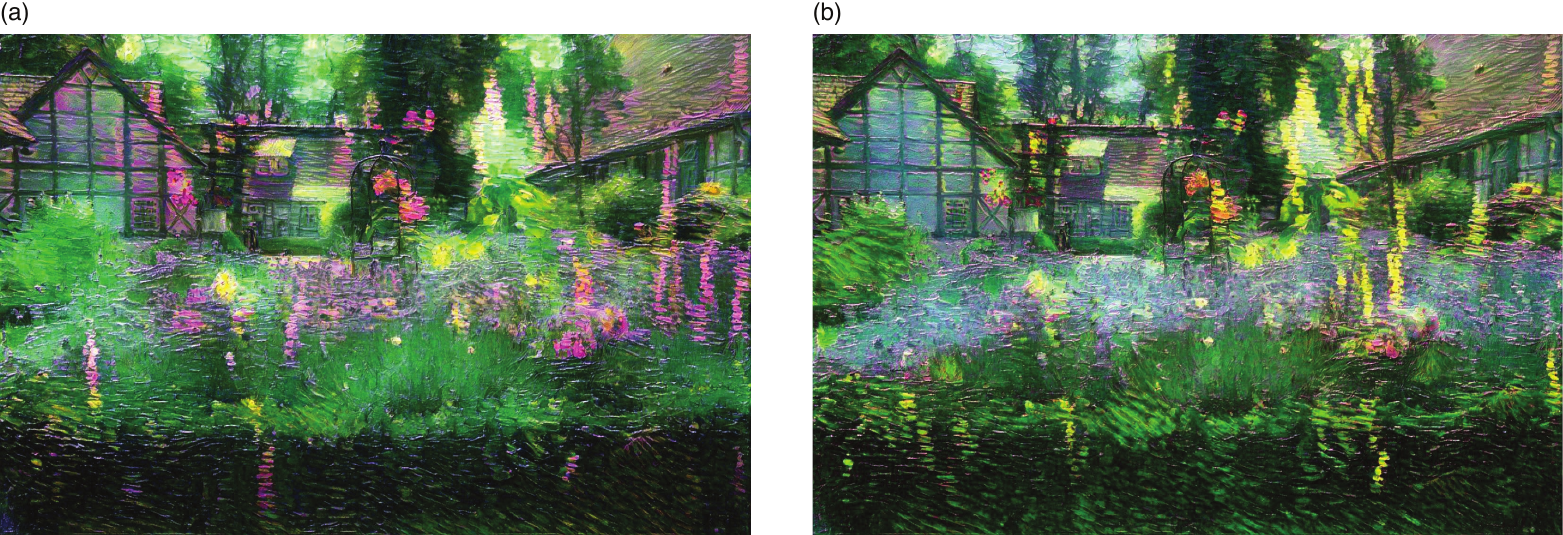}
\caption{\label{fig:color_transfer}
Comparison of linear color transfer onto the style image from Figure \ref{fig:teaser}.
\textbf{(a)} Cholesky color transfer %applied to the style image before style transfer
\textbf{(b)} Image Analogies color transfer.
The latter better preserves the colors of the source image.
%applied to the style image before style transfer
}
\end{figure}

In general, we find that results using the Image Analogies color transfer (Fig.~\ref{fig:color_transfer}(b)) look better than those using the Cholesky transfer (Fig.~\ref{fig:color_transfer}(a)). Furthermore, the Cholesky transform has the conceptually-undesirable property that it depends on the channel ordering, i.e., using RGB images will give different results from BGR.  We also experimented with the Monge-Kantorovitch linear transform \cite{mkl} that minimizes the pixel-wise $L_2$-distance to the image before color transfer, and found the results to be essentially indistinguishable from the Image Analogies approach.

In general, we find that the color matching method works reasonably well with neural style transfer. This is in contrast to Image Analogies, where it gave poor synthesis results \cite{hertzmann-phd}.

\begin{figure}
\includegraphics{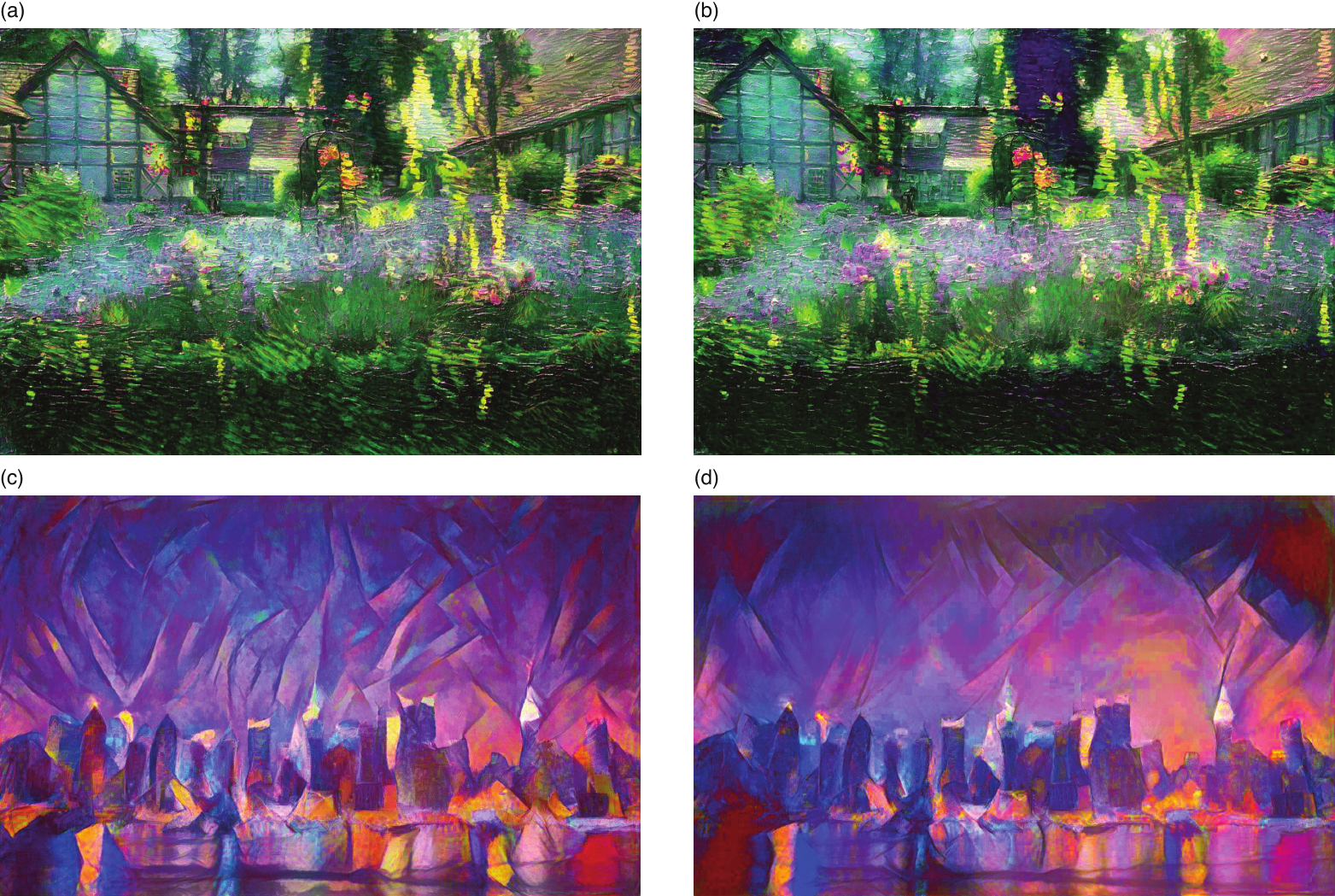}
\caption{\label{fig:pre_post}
Comparison between matching the color distribution before or after style transfer. The Image Analogies color transfer is used.
\textbf{(a)} 
Color transfer before synthesis (same as Figure \ref{fig:teaser}(d)).
%Color transfer applied to the style image from Figure \ref{fig:teaser} before style transfer.
\textbf{(b)} Color transfer after synthesis. %applied to the style image from Figure \ref{fig:teaser} after style transfer.
There are subtle differences, including purple plants in the middle-top of  (b).
\textbf{(c)} 
Color transfer before synthesis (same as Figure \ref{fig:teaser_NY}(d)).
\textbf{(d)} 
Color transfer after synthesis.
Here, the sky loses the ``cubist texture" of the style image.
}
\end{figure}

We also test transferring the color distribution to the output of neural style transfer instead of to the style image.
In other words, neural style transfer is computed from the original inputs $S$ and $C$, and then the output $T$ is color-matched to $C$, producing a new output $T'$.
%Instead of transferring the color distribution onto the style image before style transfer, we can also transfer the color distribution from the content image directly onto the output image ($T$) after running the standard style transfer from \cite{gatys} to produce a new output image $T'$. 
We find that  transferring the color histogram before style transfer leads to better results (Fig.~\ref{fig:pre_post}). This is particularly apparent for the second example in Figure \ref{fig:pre_post}, where the cubist texture is not transferred as completely to the sky. 
%Here one can see that first changing the color statistics of the style image can have a severe effect on the outcome of the style transfer. 
It appears to be easier for the algorithm to transfer the style between two images with similar color distribution. This can be explained by the reduced competition between reconstructing the content image and simultaneously matching the texture information from the style image.

\section{Approach \#2: Luminance-only transfer}

The second method we consider is to perform style transfer only in the luminance channel, as used in Image Analogies \cite{image-analogies}. This is motivated by the observation that visual perception is far more sensitive to changes in luminance than in color \cite{wandell}.

\begin{figure}
\includegraphics{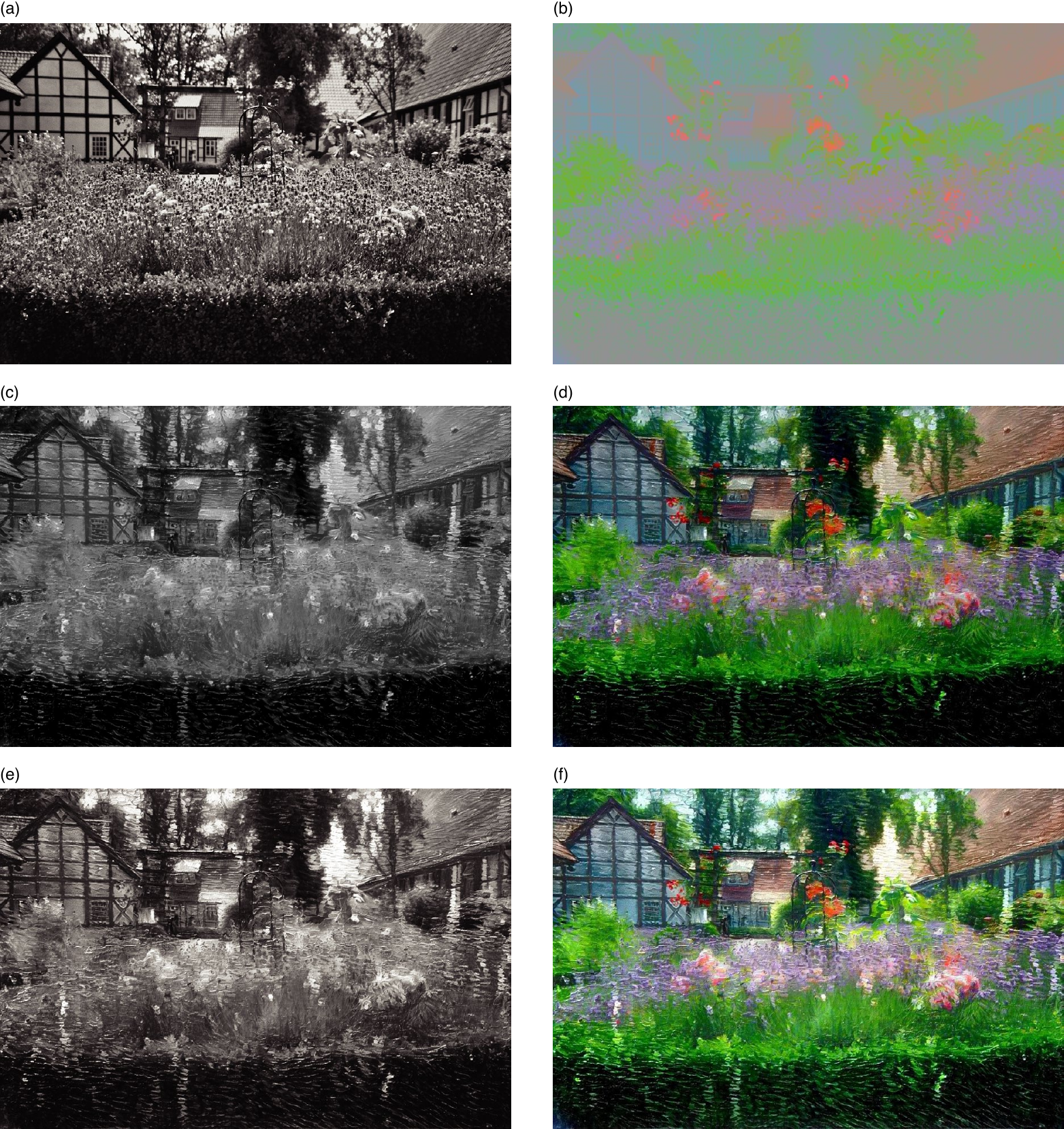}
\caption{\label{fig:luminance}
Luminance-only synthesis, on the images of Figure \ref{fig:teaser}.
\textbf{(a)} Luminance channel ($Y$) of input photograph. 
\textbf{(b)} Color channels ($I,Q$) of input photograph.
\textbf{(c)} Style transfer result in luminance channel.
\textbf{(d)} Combination of synthesized luminance and source color channels.
\textbf{(e)} 
Synthesis, using luminance-histogram matching before synthesis.
\textbf{(f)} 
Combination of color and luminance channels, using luminance-histogram matching before synthesis.
}
\end{figure}

The modification is simple. The luminance channels $L_S$ and $L_C$ are first extracted from the style and content images. 
Then the neural style transfer algorithm is applied to these images to produce an output luminance image $L_T$.
Using the YIQ color space, the color information of the content image are represented by the $I$ and $Q$ channels; these are combined with $L_T$ to produce the final color output image (Fig. \ref{fig:luminance}(c,d)).

If there is a substantial mismatch between the luminance histogram of the style and the content image, it can be helpful to match the histogram of the style luminance channel $L_S$ to that of the content image $L_C$ before transferring the style. 
As in the previous section, we use a linear map that matches the second-order statistics of the content image (Fig. \ref{fig:luminance}(e,f)).  Let $\mu_S$  and $\mu_C$ be the mean luminances of the two images, and $\sigma_S$ and $\sigma_C$ be their standard deviations. Then each luminance pixel in the style image is updated as:
\begin{equation}
L_{S'} = \frac{\sigma_C}{\sigma_S} (L_S - \mu_S) + \mu_C
\end{equation}

\section{Comparison and discussion}

Here we present two simple methods to preserve the color of the content image in the neural style transfer algorithm \cite{gatys}:
\begin{enumerate}
    \item Linear color transfer onto the style image, before style transfer.
    \item Style transfer only in the luminance channel.
\end{enumerate}
Both methods give perceptually-interesting results but have different advantages and disadvantages. 

\begin{figure}
\includegraphics{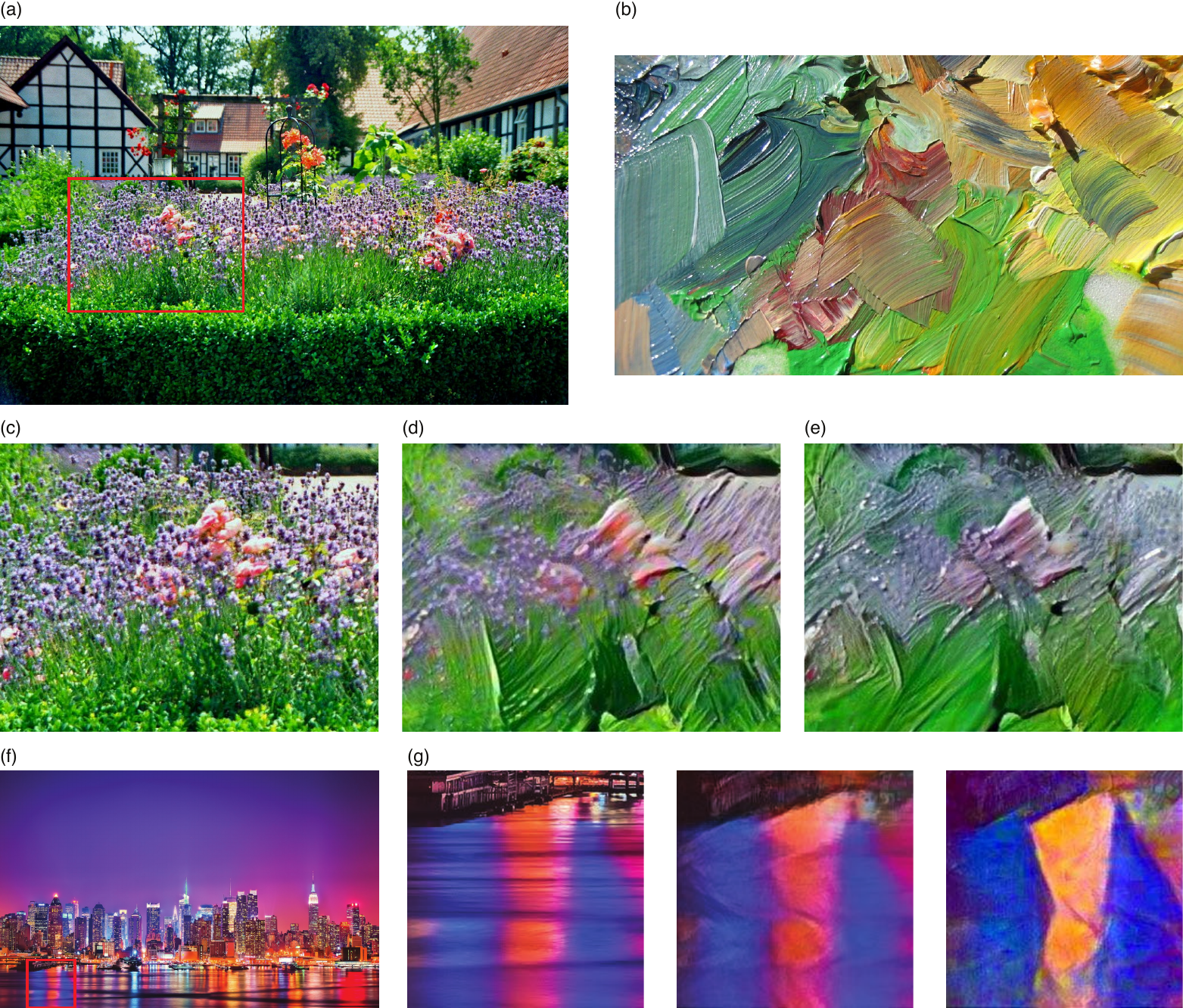}
\caption{\label{fig:discussion}
Advantages and disadvantages of the two methods presented.
\textbf{(a)} Content image from Fig. \ref{fig:teaser} with marked region.
\textbf{(b)} Style image. This example contains prominent brushstrokes.
\textbf{(c)} Detail of a region in the content image (marked in (a))
\textbf{(d)} Luminance-only synthesis for this region. Color is preserved perfectly, but dependencies between luminance and color are lost: colors do not align well with strokes.
\textbf{(e)} Color-matching synthesis for this region. Dependencies between luminance structure and colors are preserved but the colors are not preserved exactly.
\textbf{(f)} Content image of a second example (same as Figure \ref{fig:teaser_NY} (c) and (d)). Content image has a marked region in the bottom left corner.
\textbf{(g)} Same as (c),(d) and (e) for the second example. Even without brushstrokes, there is an effect. Dependencies between color and luminance structure are only preserved with the color transfer method.
}
\end{figure}

The first method is naturally limited by how well the color transfer from the content image onto the style image works. The color distribution often cannot be matched perfectly, leading to a mismatch between the colors of the output image and that of the content image (Fig.~\ref{fig:discussion}(e)). The synthesis also replicates ``content" structures from the van Gogh style scene, i.e., the pattern of reflections on the river appear as vertical yellow stripes of brushstrokes in the output.

In contrast, the second method  preserves the colors of the content image perfectly. However, dependencies between the luminance and the color channels are lost in the output image (Fig.~\ref{fig:discussion} (d)). This is particularly apparent for styles with prominent brushstrokes. In Figure \ref{fig:discussion}(d), colors are no longer aligned to strokes. That means a single brushstroke can have multiple colors, which does not happen in real paintings. In comparison, when using full style transfer and color matching, the output image really consists of strokes which are blotches of paint, not just variations of light and dark. 

One potential advantage of the luminance-based method is that it reduces the dimensionality of the optimization problem for the neural synthesis. The neural synthesis algorithm performs numerical optimization of the output image, and luminance-only synthesis has one-third fewer parameters. However, it is unclear that there is any practical advantage in typical GPU implementations.

In future work, it would be interesting to explore how the two statistical models in here (color statistics vs.~CNN activations) might be unified, and to explore more sophisticated color transfer and adjustment procedures \cite{CGF:CGF12671}.

\bibliographystyle{acm}
\bibliography{paper}

\end{document}